\documentclass[10pt,twocolumn,letterpaper]{article}

\usepackage[pagenumbers]{cvpr} % To force page numbers, e.g. for an arXiv version

\usepackage{url}            % simple URL typesetting
\usepackage{booktabs}       % professional-quality tables
\usepackage{amsfonts}       % blackboard math symbols
\usepackage{nicefrac}       % compact symbols for 1/2, etc.
\usepackage{microtype}      % microtypography
\usepackage{xcolor}         % colors
\usepackage{xspace}  
\usepackage{graphicx}
\usepackage{booktabs}
\usepackage{adjustbox}
\usepackage{multicol}
\usepackage{multirow}
\usepackage{hyphenat}
\usepackage{subcaption}
\usepackage{bm}
\usepackage{amsmath}
\usepackage{makecell}
\usepackage{pifont}
\usepackage{lipsum}
\usepackage{bbding}
\usepackage{wasysym}
\usepackage{amssymb}
\usepackage{wrapfig}
\usepackage{cuted}
\usepackage{capt-of}

\usepackage{color, colortbl}
\usepackage{physics}
\usepackage{footnote}
\usepackage{tabularx}
\usepackage{float}

\usepackage{siunitx}
\usepackage{tabularx,booktabs}
\usepackage{etoolbox}
\usepackage{algorithm}
\usepackage{listings}

\usepackage{threeparttable}
\usepackage{subcaption}
\usepackage{enumitem}

\usepackage{textpos}  % for superposing a text box
\usepackage{tikz}

\newcommand{\methodname}{Empirical Variational Autoencoder\xspace}
\newcommand{\methodnameshort}{EVA\xspace}

\DeclareRobustCommand{\gridthreefive}{%
  \raisebox{-0.5ex}{%
    \tikz[x=0.333em,y=0.333em]{
      \fill[gray!40] (1,1) rectangle (3,2);
      \fill[gray!40] (0,0) rectangle (3,1);

      \draw[line width=0.04em] (0,0) rectangle (3,3);

      \draw[line width=0.04em] (1,0) -- (1,3);
      \draw[line width=0.04em] (2,0) -- (2,3);

      \draw[line width=0.04em] (0,1) -- (3,1);
      \draw[line width=0.04em] (0,2) -- (3,2);
    }%
  }%
}

\definecolor{cvprblue}{rgb}{0.21,0.49,0.74}
\usepackage[pagebackref,breaklinks,colorlinks,allcolors=cvprblue]{hyperref}
\def\paperID{*****} % *** Enter the Paper ID here
\def\confName{CVPR}
\def\confYear{2026}

\title{Empirical Variational Autoencoder}

\author{Kaede Shiohara\\
The University of Tokyo\\
{\tt\small shiohara@cvm.t.u-tokyo.ac.jp}
}

\begin{document}
\twocolumn[{%
    \renewcommand\twocolumn[1][]{#1}%
    \maketitle
    \begin{center}
      \vspace{-4mm}
      \includegraphics[width=\textwidth]{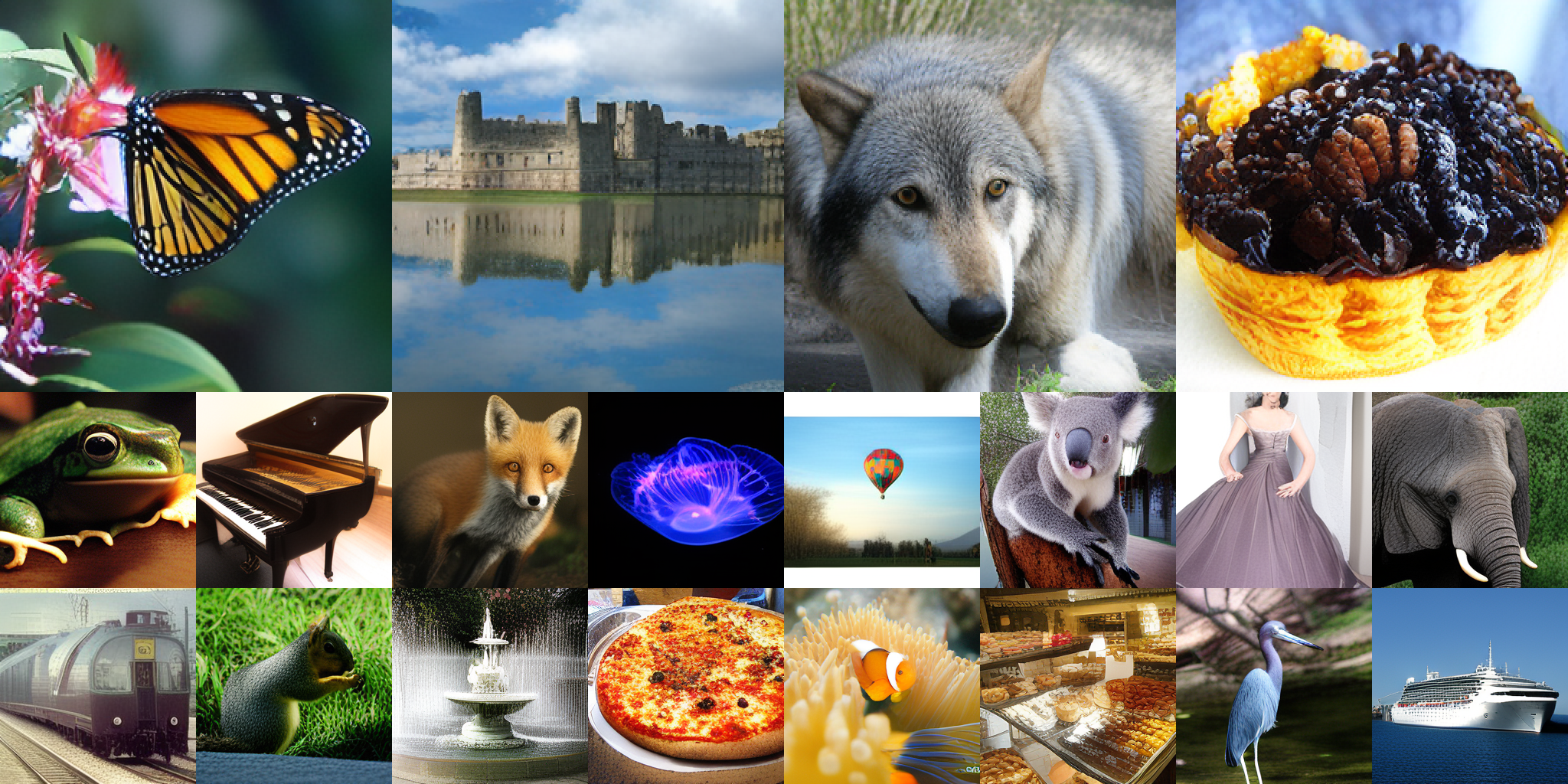}
      \captionof{figure}{Curated examples generated by \methodnameshort-L trained on ImageNet (256$\times$256).
      }
      \label{fig:teaser}
    \end{center}
}]

\begin{abstract}
We present \methodname (\methodnameshort), a general generative framework for continuous-valued (\ie, non-vector-quantized) sequences.
\methodnameshort is based on the evidence lower bound of the Variational Autoencoder (VAE) but learns autoregressive latent priors empirically from training data, which can be implemented only by an additional single linear layer on top of VAEs. 
By replacing the conventional standard-Gaussian constraint with the self-predicted priors, EVA significantly alleviates the latent distribution gap between prior and posterior which is typically observed in conventional VAEs, and leads to high-fidelity ancestral sampling for sequential data generation.
Extensive experiments on image and sound synthesis demonstrate that \methodnameshort achieves competitive generation quality with autoregressive diffusion baselines despite its much faster inference time.
Project page:~\url{https://mapooon.github.io/EVAPage}.
\end{abstract}

\section{Introduction}
Autoregressive (AR) models are a dominant approach to sequence generation, but their standard formulation is most natural for discrete-valued data, where each predictive distribution is categorical. 
For continuous-valued sequences, direct regression losses such as mean squared error do not model the full next-token density and therefore struggle to capture multimodality and uncertainty. 
Existing continuous AR approaches address this issue by employing continuous generative heads such as Gaussian mixture model~\cite{givt} and diffusion~\cite{mar}, but these methods either impose fixed parametric families or incur substantial inference costs.
Variational autoencoders (VAEs)~\cite{vae} provide a theoretically well-supported generative framework for continuous data through latent-variable modeling. 
However, conventional VAEs can suffer from a mismatch between the dependencies remaining across latent positions in the posterior and the factorized standard-Gaussian prior used for generation, which significantly degrades sampling quality. 
In practice, this limitation has contributed to a trend in which VAE-style models are more often used as latent compressors or tokenizers than as standalone generators in modern large-scale generative pipelines~\cite{vqvae,rombach2022high}.

In this work, \textbf{we revisit the role of VAEs as generative models} especially for sequential data.
Our proposed \methodname (\methodnameshort) is built on the evidence lower bound (ELBO) of VAEs, but replaces the traditional fixed latent prior with an autoregressive prior learned empirically from data.
This simple modification significantly aligns the prior distribution with the posterior and enables high-fidelity sequential sampling. 
Experiments on a controlled toy benchmark, ImageNet~\cite{russakovsky2015imagenet}, and VGGSound~\cite{chen2020vggsound} show that \methodnameshort achieves competitive results with autoregressive generators~\cite{givt,mar} while maintaining substantially more efficient ancestral inference than diffusion-based AR approaches.

\section{Related Work}
\noindent\textbf{AR Models in Discrete-valued Space.} 
Autoregressive (AR) models factorize the joint distribution as $p(x)=\prod_t p(x_t\mid x_{<t})$ and have been highly successful on discrete sequential data, from early RNN-based models~\cite{10.5555/3104482.3104610,graves2013generating} to modern Transformer-based models such as GPT~\cite{gpt2,gpt3,attention}. 
For continuous signals such as images and audio, a common strategy is to discretize the data or map them to learned discrete codes, enabling next-token prediction in pixel space~\cite{pixelrnn,pixelcnn,pixelcnnpp}, waveform space~\cite{wavenet}, or latent token space~\cite{vqvae,taming,var}. 
However, such discretization often introduces quantization error and optimization difficulties due to non-differentiable code assignment~\cite{mentzer2024finite}.
In contrast to discrete AR models, our method operates directly on continuous-valued tokens.

\noindent\textbf{AR Models in Continuous-valued Space.}
Beyond discretization-based approaches, another line of work models continuous-valued sequences directly in the original space.
Early approaches combine Gaussian mixture models~\cite{graves2013generating,givt} with AR models, but their expressiveness is limited by a pre-defined, fixed number of mixture components. 
More recent approaches introduce implicit objectives, such as diffusion-based autoregressive models~\cite{mar} or likelihood-free objectives based on energy scores~\cite{shao2025continuous}, trading off either inference efficiency or mode coverage. 
In contrast, our method models each conditional distribution as a continuous mixture induced by latent marginalization, while learning an autoregressive prior for high-quality ancestral sampling.

\noindent\textbf{Challenges in VAEs as Generative Models.}
Early VAEs~\cite{vae} typically represent an entire observation using a single global latent variable, resulting in an inherent trade-off between reconstruction fidelity and prior regularization. 
To improve this formulation, more expressive priors are introduced, including VampPrior~\cite{vampprior}, resampled priors~\cite{lars}, and flow-based priors~\cite{ding-gimpel-2021-flowprior}, while some studies tried to combine VAEs with AR decoders~\cite{bowman2016generating,pixelvae,deltavae}. 
However, as autoregressive decoders become more expressive, they can increasingly ignore the latent variable, leading to the posterior collapse issue. 
This creates a tension between maintaining latent-variable usage and scaling increasingly expressive decoders.
Another direction increases latent capacity by distributing latent variables over spatial or hierarchical representations, rather than the single global latent variable. 
Deep hierarchical VAEs such as NVAE~\cite{vahdat2020NVAE} and VDVAE~\cite{child2020very} model dependencies across hierarchical latent groups, where each group is typically a spatial latent feature map sampled in parallel, but use factorized conditional distributions within each group.
This factorization leaves dependencies within each spatial latent map unmodeled, and prior--aggregate posterior mismatch has been reported to persist even in deep hierarchical VAEs~\cite{aneja2021contrastive}. 
In contrast, EVA autoregressively conditions each latent token on all preceding latent positions, giving the prior the capacity to represent dependencies across the entire preceding latent sequence and enabling high-fidelity ancestral sampling.

\section{Proposed Method}
We present \methodname (\methodnameshort), a general generative framework for continuous-valued sequences.
\methodnameshort is based on the evidence lower bound of Variational Autoencoder (VAE)~\cite{vae} but replaces the fixed standard-Gaussian constraint with a self-predicted autoregressive prior.
This allows structured dependencies to remain in the latent representation while adapting the prior to the posterior, enabling ancestral sampling from a prior aligned with the learned latent structure.
We start by revisiting the theory of VAE as follows.

\begin{figure*}[t]
    \centering
    \includegraphics[width=1.0\linewidth]{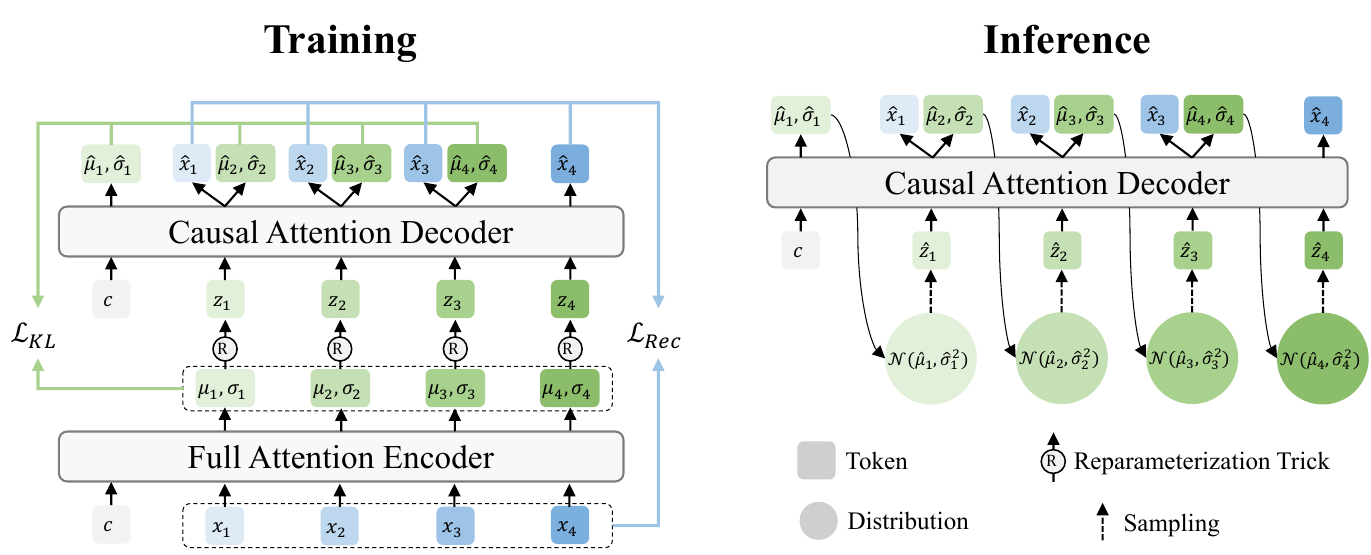}
    \caption{\textbf{Framework of \methodnameshort.} 
    During training, the encoder maps the observed sequence $x_{1:T}$ to latent variables $z_{1:T}$, the decoder reconstructs each token $\hat{x}_t$ from the latent $z_{1:t}$, and learns the autoregressive latent prior $p(z_t \mid z_{1:t-1})$ by matching it to the posterior $q(z_t \mid x_{1:T})$.
    During inference, the model autoregressively samples the latent sequence from the empirical prior and decodes each token causally to generate a continuous-valued sequence.
    }
    \label{fig:framework}
\end{figure*}

\subsection{Revisiting Variational Autoencoder}
\label{sec:vae}
Variational Autoencoder (VAE)~\cite{vae} is a latent-variable generative model that introduces a continuous latent variable $\mathbf{z}$ for observed data $\mathbf{x}$. 
It assumes the following generative process:
$p(\mathbf{x}, \mathbf{z}) = p(\mathbf{x} \mid \mathbf{z}) p(\mathbf{z})$,
where $p(\mathbf{z})$ is a prior distribution over the latent variable.
The learning objective is to maximize the marginal log-likelihood:
$\log p(\mathbf{x}) = \log \int p(\mathbf{x}, \mathbf{z}) \, d\mathbf{z}$.
Since the true posterior $p(\mathbf{z} \mid \mathbf{x})$ is generally intractable, we introduce an approximate posterior $q(\mathbf{z} \mid \mathbf{x})$. 
Consider the KL divergence between these distributions using Bayes' rule:
\begin{align}
&D_{\mathrm{KL}}\!\left(
q(\mathbf{z} \mid \mathbf{x})
\;\|\;
p(\mathbf{z} \mid \mathbf{x})
\right)
=
\mathbb{E}_{q(\mathbf{z} \mid \mathbf{x})}
\left[
\log \frac{q(\mathbf{z} \mid \mathbf{x})}{p(\mathbf{z} \mid \mathbf{x})}
\right] \\
&=
\mathbb{E}_{q(\mathbf{z} \mid \mathbf{x})}
\left[
\log \frac{q(\mathbf{z} \mid \mathbf{x}) p(\mathbf{x})}{p(\mathbf{x}, \mathbf{z})}
\right] \\
&=\mathbb{E}_{q(\mathbf{z} \mid \mathbf{x})}
\left[
\log q(\mathbf{z} \mid \mathbf{x})
-
\log p(\mathbf{x}, \mathbf{z})
\right]
+
\log p(\mathbf{x}).
\end{align}
Rearranging terms gives
\begin{align}
\log p(\mathbf{x})
&=
\mathbb{E}_{q(\mathbf{z}\mid\mathbf{x})}
\left[
\log p(\mathbf{x},\mathbf{z})
-
\log q(\mathbf{z}\mid\mathbf{x})
\right]
\nonumber\\
&\quad\quad\quad+
D_{\mathrm{KL}}\!\left(
q(\mathbf{z}\mid\mathbf{x})
\,\|\,p(\mathbf{z}\mid\mathbf{x})
\right)
\\
&=
\underbrace{
\mathbb{E}_{q(\mathbf{z}\mid\mathbf{x})}
\left[
\log p(\mathbf{x}\mid\mathbf{z})
\right]
-
D_{\mathrm{KL}}\!\left(
q(\mathbf{z}\mid\mathbf{x})
\,\|\,p(\mathbf{z})
\right)
}_{\mathrm{ELBO}(\mathbf{x})}
\nonumber\\
&\quad\quad\quad+
\underbrace{
D_{\mathrm{KL}}\!\left(
q(\mathbf{z}\mid\mathbf{x})
\,\|\,p(\mathbf{z}\mid\mathbf{x})
\right)
}_{\ge 0}
\\
&\ge
\mathrm{ELBO}(\mathbf{x}).
\end{align}
where ${\mathrm{ELBO}}(\mathbf{x})$ denotes the \textit{\textbf{e}vidence \textbf{l}ower \textbf{bo}und}, which serves as a tractable lower bound of the marginal log-likelihood $\log p(\mathbf{x})$.
The ELBO is optimized by parameterizing the conditional distribution and the approximate posterior with neural networks, denoted as $p_{\theta}(\mathbf{x} \mid \mathbf{z})$ and $q_{\phi}(\mathbf{z} \mid \mathbf{x})$, respectively. 
A common choice is to model the approximate posterior as a diagonal Gaussian:
\begin{equation}
q_{\phi}(\mathbf{z} \mid \mathbf{x})
=
\mathcal{N}
\left(
\mathbf{z};
\boldsymbol{\mu}_{\phi}(\mathbf{x}),
\boldsymbol{\sigma}_{\phi}^2(\mathbf{x})
\right),
\end{equation}
and to employ the reparameterization trick for gradient-based optimization.
\begin{equation}
\mathbf{z}
=
\boldsymbol{\mu}_{\phi}(\mathbf{x})
+
\boldsymbol{\sigma}_{\phi}(\mathbf{x}) \odot \boldsymbol{\epsilon},
\quad
\boldsymbol{\epsilon} \sim \mathcal{N}(\mathbf{0}, \mathbf{1}).
\end{equation}

Accordingly, the training objective is to minimize the negative ELBO, or VAE loss:
\begin{align}
&\mathcal{L}_{\mathrm{VAE}}(\mathbf{x}; \theta, \phi)
= -{\mathrm{ELBO}}(\mathbf{x}; \theta, \phi)
\nonumber\\
&= D_{\mathrm{KL}}
\left(
q_{\phi}(\mathbf{z} \mid \mathbf{x})
\;\|\;
p(\mathbf{z})
\right)
-
\mathbb{E}_{q_{\phi}(\mathbf{z} \mid \mathbf{x})}
\left[
\log p_{\theta}(\mathbf{x} \mid \mathbf{z})
\right],
\end{align}
where the prior over the latent variable is typically chosen as a standard normal distribution $p(\mathbf{z}) = \mathcal{N}(\mathbf{0}, \mathbf{1})$.
This conventional choice induces a trade-off between the two objectives in the ELBO: optimizing the reconstruction term encourages the posterior to capture detailed data-dependent structure, thereby widening its gap from the fixed standard-Gaussian prior, whereas optimizing the KL term reduces this gap by forcing the posterior toward the prior, often at the expense of reconstruction quality. 
This makes it difficult for traditional VAEs to simultaneously obtain accurate reconstructions and high-quality prior sampling.

\subsection{\methodname (\methodnameshort)}
We start by considering a case where $\mathbf{x}$ and $\mathbf{z}$ are sequences with a length of $T$ as denoted:
\begin{equation}
\mathbf{x} = x_{1:T} = \{x_1,x_2,...,x_{T}\} \in \mathbb{R}^{T\times D} \quad\text{where}\, x_{t} \in \mathbb{R}^{D},
\end{equation}
\begin{equation}
\mathbf{z} = z_{1:T} =\{z_1,z_2,...,z_{T}\} \in \mathbb{R}^{T\times C} \quad\text{where}\, z_{t} \in \mathbb{R}^{C}.
\end{equation}
We assume a causal decoder, an autoregressive prior, and a mean-field posterior:
{\small
\begin{align}
p(\mathbf{x} \mid \mathbf{z})
=
\prod_{t=1}^{T} p(x_t \mid z_{1:t}), \\
p(\mathbf{z})
=
\prod_{t=1}^{T} p(z_t \mid z_{1:t-1}), \\
q(\mathbf{z} \mid \mathbf{x})
=
\prod_{t=1}^{T} q(z_t \mid \mathbf{x}),
\end{align}
}
where $p(z_1 \mid z_{1:0}) := p(z_1)$.
Under these assumptions, the VAE ELBO is reformulated as:
\begin{align}
&{\mathrm{ELBO}}(\mathbf{x})
=
\mathbb{E}_{q(\mathbf{z} \mid \mathbf{x})}
\left[
\log p(\mathbf{x} \mid \mathbf{z})
\right]
-
D_{\mathrm{KL}}
\left(
q(\mathbf{z} \mid \mathbf{x})
\;\|\;
p(\mathbf{z})
\right)
\nonumber\\
&=
\sum_{t=1}^{T}
\Big(
\mathbb{E}_{q(z_{1:t} \mid \mathbf{x})}
\left[
\log p(x_t \mid z_{1:t})
\right]
\nonumber\\
&-
\mathbb{E}_{q(z_{1:t-1} \mid \mathbf{x})}
\left[
D_{\mathrm{KL}}
\left(
q(z_t \mid \mathbf{x})
\;\|\;
p(z_t \mid z_{1:t-1})
\right)
\right]
\Big),
\end{align}
where $p(z_1 \mid z_{1:0})=p(z_1)$.
We call this \textbf{\methodnameshort ELBO} to distinguish it from the VAE ELBO.

Then, we parameterize the conditional distributions as an encoder $q_{\phi}(z_t \mid \mathbf{x})$ and a causal decoder $p_{\theta}(x_t \mid z_{1:t})$ to optimize \methodnameshort ELBO.
Moreover, \textit{most importantly}, \methodnameshort optimizes latent priors empirically from training data by an additional network, called \textit{predictor}, $p_{\omega}(z_t \mid z_{1:t-1})$,
unlike the original VAE that constrains the priors simply to the standard Gaussian $\mathcal{N}(\mathbf{0},\mathbf{1})$ shared across different spatial (or temporal) indices,
as formulated:
\begin{equation}
p_{\omega}(z_t \mid z_{1:t-1})
=
\mathcal{N}
\left(
z_t;
\hat{\mu}_t,
\hat{\sigma}_t^2
\right),
\end{equation}
where $\hat{\mu}_t = \mu_{\omega}(z_{1:t-1})$ and $\hat{\sigma}_t = \sigma_{\omega}(z_{1:t-1})$.
Finally, the training objective for \methodnameshort is obtained as the negative \methodnameshort ELBO, denoted as the \methodnameshort loss:
\begin{align}
\label{eq:eva-loss}
&\mathcal{L}_{\mathrm{\methodnameshort}}(\mathbf{x}; \theta, \phi, \omega)
= -{\mathrm{ELBO}}(\mathbf{x}; \theta, \phi, \omega) 
\nonumber\\
&= \sum_{t=1}^{T}
\Big(
\mathbb{E}_{q_{\phi}(z_{1:t-1} \mid \mathbf{x})}
\left[
D_{\mathrm{KL}}
\left(
q_{\phi}(z_t \mid \mathbf{x})
\;\|\;
p_{\omega}(z_t \mid z_{1:t-1})
\right)
\right]
\nonumber\\
&\quad-
\mathbb{E}_{q_{\phi}(z_{1:t} \mid \mathbf{x})}
\left[
\log p_{\theta}(x_t \mid z_{1:t})
\right]
\Big),
\end{align}
where $p_\omega(z_1 \mid z_{1:0})$ is simply defined as $p_\omega(z_1)$.

\begin{figure*}[t]
    \centering
    \includegraphics[width=1.0\linewidth]{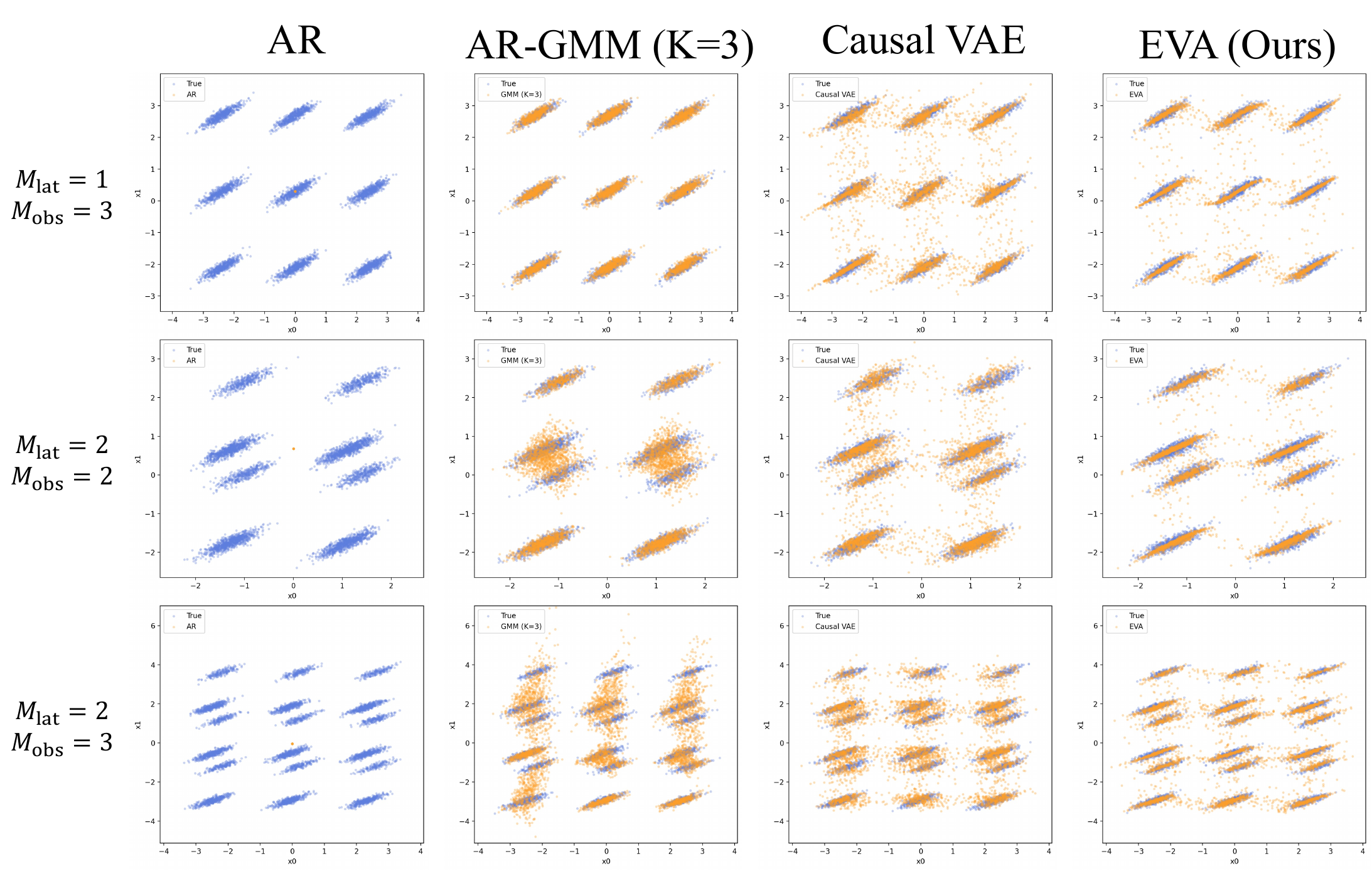}
    \caption{\textbf{Toy Experiment on Multimodal Sequence Structure.} 
    Each panel shows samples in the two-dimensional $(x_0, x_1)$ space, with the horizontal axis corresponding to $x_0$ and the vertical axis corresponding to $x_1$.
    The {\color{blue}blue} points are sampled from the true data distribution while the {\color{orange}orange} points are generated by each model.
    AR tends to average over plausible modes and fails to recover the multimodal joint structure.
    GMM fails once the multimodal structure exceeds its finite mixture capacity.
    Causal VAE captures the major modes, but still produces noticeable inter-mode samples.
    In contrast, \methodnameshort concentrates samples on the true modes and substantially suppresses inter-mode samples, indicating that the learned autoregressive prior better represents latent dependencies across sequence steps.
    Best viewed in zoom.
    }
    \label{fig:toy}
\end{figure*}

\noindent\textbf{Sampling.} 
At inference, generation is performed ancestrally in the latent space: we first sample $z_1 \sim p_\omega(z_1)$, and then sample $z_t \sim p_{\omega}(z_t \mid z_{1:t-1})$ for $t=2,\ldots,T$. The output is decoded causally as $x_t \sim p_{\theta}(x_t \mid z_{1:t})$.

\noindent\textbf{Conditional Generation.}
Given a condition vector $c$, \methodnameshort is extended to conditional generation:
\begin{equation}
p(\mathbf{x}, \mathbf{z} \mid c)
=
p(\mathbf{x} \mid \mathbf{z}, c)\, p(\mathbf{z} \mid c).
\label{eq:cond_joint}
\end{equation}
Using a causal conditional decoder, an autoregressive conditional prior, and a mean-field conditional posterior, substituting these factorizations into the standard conditional ELBO yields
{\small
\begin{align}
&{\mathrm{ELBO}}(\mathbf{x}, c)
=
\sum_{t=1}^{T}
\Big(
\mathbb{E}_{q(z_{1:t} \mid \mathbf{x}, c)}
\left[
\log p(x_t \mid z_{1:t}, c)
\right] 
\nonumber\\
&\quad-
\mathbb{E}_{q(z_{1:t-1} \mid \mathbf{x}, c)}
\left[
D_{\mathrm{KL}}
\left(
q(z_t \mid \mathbf{x}, c)
\;\|\;
p(z_t \mid z_{1:t-1}, c)
\right)
\right]
\Big),
\label{eq:cond_elbo}
\end{align}
}
where $p(z_1 \mid z_{1:0}, c):=p(z_1 \mid c)$.
We parameterize these distributions by a conditional encoder
$q_{\phi}(z_t \mid \mathbf{x}, c)$, a conditional decoder
$p_{\theta}(x_t \mid z_{1:t}, c)$, and a conditional predictor
$p_{\omega}(z_t \mid z_{1:t-1}, c)$.
In particular, the predictor defines a conditional Gaussian prior:
\begin{align}
p_{\omega}(z_t \mid z_{1:t-1}, c)
&=
\mathcal{N}
\left(
z_t;
\hat{\mu}_t,
\hat{\sigma}_t^2
\right),
\\
\hat{\mu}_t&=\mu_{\omega}(z_{1:t-1}, c),
\\
\hat{\sigma}_t&=\sigma_{\omega}(z_{1:t-1}, c).
\label{eq:cond_predictor}
\end{align}
We optimize the conditional objective
\begin{equation}
\mathcal{L}_{\mathrm{\methodnameshort}}(\mathbf{x}, c; \theta, \phi, \omega)
=
-{\mathrm{ELBO}}(\mathbf{x}, c; \theta, \phi, \omega).
\label{eq:cond_loss}
\end{equation}
At inference, we sample $z_1 \sim p_{\omega}(z_1 \mid c)$ and
$z_t \sim p_{\omega}(z_t \mid z_{1:t-1}, c)$ for $t=2,\ldots,T$,
and decode each token causally via $p_{\theta}(x_t \mid z_{1:t}, c)$.
The training and inference pipelines are illustrated in Fig.~\ref{fig:framework}.

\subsection{Multimodal Sequence Structure}
\label{sec:toy}
To understand how \methodnameshort differs from previous generative families in capturing multimodal dependencies in sequences, we study a toy experiment on the joint distribution $p(x_0, x_1)$ of two sequential observations $x_0$ and $x_1$. 
We model real-world sequential data by a minimal latent-state generative process, in which observations are generated from latent states and the latent dynamics follow a conditional transition of the form $p(z_1 \mid z_0)$.
Multimodality can arise both from branching latent transitions, which produce multiple plausible futures, and from multiple observable realizations of each latent state.

\noindent\textbf{Generative Families.}
We first compare how different generative model families represent the predictive distribution from a theoretical perspective:

\noindent 1) \textit{When an AR decoder is trained simply with the MSE loss}, the model is interpreted to estimate the mean of a \textbf{single} modal Gaussian:
\begin{equation}
p(x_t \mid x_{1:t-1}) = \mathcal{N}(x_t; \mu(x_{1:t-1}), \sigma^2 I).
\end{equation}
Under squared-error training, the optimal predictor is the conditional mean $ $ which averages over multiple plausible outcomes when the conditional distribution is multimodal.

\noindent 2) \textit{When an AR decoder is equipped with a GMM head}, the predictive distribution is formulated as
\begin{equation}
p(x_t \mid x_{1:t-1}) = \sum_{k=1}^K \pi_k(x_{1:t-1}) \, \mathcal{N}(x_t; \mu_k(x_{1:t-1}), \sigma_k(x_{1:t-1})).
\end{equation}
The limitation of the GMM head is that it represents $p(x_t \mid x_{1:t-1})$ as a finite mixture with a pre-defined number of components $K$.
In this toy experiment, we consider $K=3$.

\noindent 3) \textit{VAE equipped with a causal decoder} represents $p(x_t \mid x_{1:t-1})$ through latent marginalization:
\begin{equation}
\label{eq:vaepredictive}
p(x_t \mid x_{1:t-1})
=
\int \mathcal{N}(x_t; \mu(z_{1:t}), \sigma^2 I)\, p(z_{1:t} \mid x_{1:t-1})\, dz_{1:t}.
\end{equation}
Since $z_{1:t}$ is continuous, this marginalization induces a continuous mixture of Gaussian components, allowing the predictive distribution to be multimodal.

4) \textit{Our \methodnameshort} uses the same latent-variable predictive family as VAE (Eq.~\ref{eq:vaepredictive}), but it replaces the fixed prior $\mathcal{N}(\mathbf{0},\mathbf{1})$ of VAE with an empirical autoregressive prior $p_\omega(z_t \mid z_{1:t-1})$. 
Therefore, the key difference from causal VAE is whether the prior represents dependencies across latent positions $t$, which in turn allows latent marginalization to preserve multimodal predictive structure.

\noindent\textbf{Toy Setting.}
We construct a synthetic process on a one-dimensional latent sequence and one-dimensional observations. 
First, we sample $z_0 \sim \mathcal{N}(0, \sigma_0^2)$ and then $z_1$ is generated from $z_0$ through a state-dependent branching transition as
$m \sim p(m \mid z_0)$ and $z_1 = a z_0 + b + c_m + \varepsilon_z$,
where $m \in \{1,\ldots,M_{\mathrm{lat}}\}$ is a discrete branch variable, $M_{\mathrm{lat}}$ denotes the number of latent transition modes, $a$ and $b$ are fixed scalar coefficients, $c_m$ is a branch-specific latent offset, and $\varepsilon_z$ is Gaussian noise.
Observations are then generated via multimodal offsets:
$x_0 = z_0 + o_{u_0} + \varepsilon_0$ and $x_1 = z_1 + o_{u_1} + \varepsilon_1$,
where $u_0, u_1 \in \{1,\ldots,M_{\mathrm{obs}}\}$ are discrete observation-mode indices, $M_{\mathrm{obs}}$ denotes the number of observable modes per latent state, $o_{u_0}$ and $o_{u_1}$ are the corresponding observation offsets, and $\varepsilon_0,\varepsilon_1$ are Gaussian noise terms.
Thus, $M_{\mathrm{lat}}$ controls the branching complexity of the latent transition, while $M_{\mathrm{obs}}$ controls the visible multimodality in the observation space.
We set $(M_{\mathrm{lat}},M_{\mathrm{obs}})$ to $(1,3)$, $(2,2)$, or $(2,3)$.
All the hyperparameters and training details can be found in Appendix~\ref{sec:toy-detail}.

\begin{table*}[t]
  \centering
\begin{minipage}[c]{0.645\textwidth}
    \centering
    \centering 
    \begin{adjustbox}{width=1.0\linewidth}
    \begin{tabular}{l|c c | c c c | c c}
    & \multicolumn{5}{c|}{ImageNet} &\multicolumn{2}{c}{VGGSound} \\
    \cmidrule(lr){2-6} \cmidrule(lr){7-8} 
    & \multicolumn{2}{c|}{\scriptsize{\#Params (M)}} & & & & & \\
    \multicolumn{1}{l|}{Method} & Train & Infer & {FID$\downarrow$} & {IS$\uparrow$} & {Time$\downarrow$} &  {FID$\downarrow$} & {IS$\uparrow$}  \\
    \midrule
    AR &170.9 &170.9 & 200.50 & 3.60 &1.91 & 4.51 & 4.45 \\
    \midrule
    AR-GMM & 171.3 &171.3 & 21.71 & 124.23 &2.10 & 3.05 & 6.30\\
    \midrule
    AR-Diff. (12 blocks) &236.2 &236.2& 29.96 & 51.09 & 23.46 & 0.57 & 26.59\\
    AR-Diff. (24 blocks)&321.3 &321.3 & 5.44 & 262.80 &24.76 &0.54 &26.96 \\
    \midrule
    Causal VAE &171.5 &86.0 & 100.27 & 13.13 & 0.97 & 1.84 & 7.45 \\
    \midrule
    Causal VAE w/ NF & 259.9 &174.6 & 12.43 & 215.04 &1.15 & 0.65 &28.95 \\
    \midrule
    \methodnameshort & 171.8 & 86.4  & 7.68 & 178.60& 1.00 & 0.64 & 25.26 \\
    \end{tabular}
    \end{adjustbox}
    
  \end{minipage}
  \hfill
  \begin{minipage}[c]{0.345\textwidth}
    \centering

    \includegraphics[width=\linewidth]{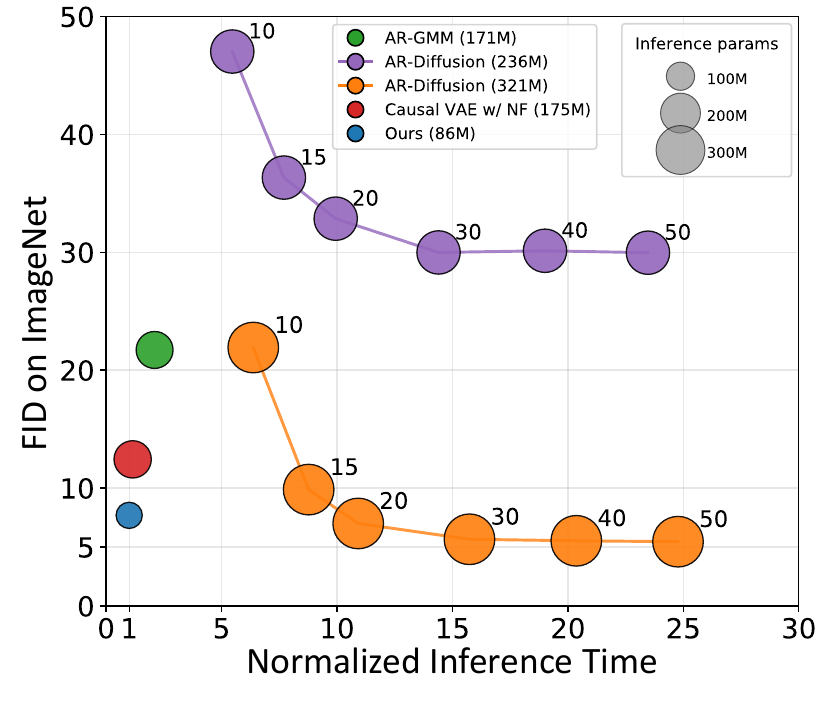}
  \end{minipage}
    \caption{\textbf{Comparison with Baselines on ImageNet (256$\times$256) and VGGSound.} 
    AR-Diffusion (12 blocks) has the same number of Transformer blocks as EVA \textit{at inference time} while AR-Diffusion (24 blocks) has the same number of Transformer blocks \textit{at training time}.
    In the right, we visualize AR-Diffusion with different denoising steps $\{10,15,20,30,40,50\}$.
    Despite the fewer inference parameters and faster inference time, our method achieves competitive generation fidelity.}
    \label{tab:result}
\end{table*}

\noindent\textbf{Results.}
Fig.~\ref{fig:toy} shows that the four model families exhibit clearly different behaviors on multimodal sequence generation.
Each sample is visualized in the two-dimensional $(x_0, x_1)$ space, where the horizontal and vertical axes correspond to $x_0$ and $x_1$, respectively.
As shown in the figure, AR trained with the MSE loss falls into the averaging trap and fails to recover the multimodal joint structure.
GMM head improves over AR and can represent $(M_{\mathrm{lat}}, M_{\mathrm{obs}})=(1,3)$ with $K=3$, since both $x_0$ and $x_1$ have three visible modes $(\leq K)$. 
However, when $(M_{\mathrm{lat}}, M_{\mathrm{obs}})=(2,2)$, $x_0$ has two modes whereas $x_1$ has four, because each of the two latent branches is further split into two observation modes. 
Since this exceeds the fixed mixture capacity $K=3$, GMM is forced to cover multiple modes with a single component. 
The same limitation is even more pronounced for $(M_{\mathrm{lat}},M_{\mathrm{obs}})=(2,3)$.
Causal VAE captures the major clusters in all settings, confirming the flexibility of latent marginalization, but it still generates a noticeable amount of inter-mode samples.
In contrast, \methodnameshort produces much fewer samples between clusters and places its mass more tightly on the true modes.

\section{Experiments}
We here validate our method on two different generation tasks including image generation on ImageNet~\cite{russakovsky2015imagenet} and sound generation on VGGSound~\cite{chen2020vggsound}.
The generated images can be seen in Fig.~\ref{fig:teaser} and Appendix.
The generated sounds are included in the project page.

\subsection{Setup.}

\noindent\textbf{Implementation.}
We implement \methodnameshort on PyTorch~\cite{paszke2019pytorch}.
We use mean squared error (MSE) as a reconstruction loss.

The encoder and decoder consist of full-attention and causal Transformer layers, respectively.
Importantly, we found in practice that the latent prior predictor $p_\omega(z_t \mid z_{1:t-1})$ can share the backbone with the decoder $p_\theta(x_t \mid z_{1:t})$;
therefore, we employ two different linear heads to predict $z_t$ and $x_t$ on top of causal Transformer backbone as illustrated in Fig.~\ref{fig:framework}.

Our model can benefit from classifier-free guidance (CFG)~\cite{cfg} as well as previous models.
Because \methodnameshort is a latent-AR model that represents $p(z_t \mid z_{1:t-1})$ rather than observation-AR models that represent $p(x_t \mid x_{1:t-1})$,
we apply CFG to the mean of the predicted latent distribution:
\begin{equation}
\mu_t(z_{1:t-1},c) \leftarrow \mu_t(z_{1:t-1}) + s\cdot (\mu_t(z_{1:t-1},c) - \mu_t(z_{1:t-1})),
\end{equation}
where $s$ is a scalar to control the strength of conditions.
Unless otherwise stated, we use $s=7$.

\noindent\textbf{Datasets and Tokenizers.}
For image synthesis, we follow the traditional ImageNet~\cite{russakovsky2015imagenet} benchmarking widely adopted in previous work~\cite{givt,mar,var} with a resolution of 256$\times$256.
The images are tokenized into 16$\times$16 tokens with 16 channels by a pre-trained VAE from~\cite{rombach2022high}.
For sound synthesis, we adopt VGGSound~\cite{chen2020vggsound} dataset that includes 309 classes\footnote{This dataset is currently provided only by third-parties and lacks a part of the original dataset.}.
The waveforms are trimmed into nine seconds and tokenized into 215 tokens with 32 channels by a pre-trained VAE from \cite{evans2025stable}.
Models are trained on the tokenized sequences and the final outputs are obtained using the pretrained VAE decoders.

\noindent\textbf{Models.}
We refer to previous generative approaches:
\textbf{AR} is a traditional AR Transformer model which is trained with MSE loss.
\textbf{AR-GMM} is a variant of AR Transformer where a Gaussian mixture model (GMM) is built on top of the Transformer to represent multi-modal density, as explored in GIVT~\cite{givt}.
\textbf{AR-Diffusion} employs a diffusion MLP head~\cite{mar} to learn implicit density.
\textbf{Causal VAE} is a VAE~\cite{vae} for sequential data by adopting a causal decoder that accesses only preceding information, which performs as our baseline model. 
\textbf{\methodnameshort just adds a single linear layer} on top of Causal VAE to predict the latent prior.
\textbf{Causal VAE w/ NF} is another baseline that learns prior by normalizing flow~\cite{zhao2025simflow}.

For fair comparisons, all the models including \methodnameshort have the same size of Transformer backbone (170M parameters consisting of 24 blocks with a hidden width of 768), excluding method-specific designs, \eg, the prior prediction head for \methodnameshort and diffusion MLP head for AR-Diffusion.
We set the latent size to 128 for Causal VAE, Causal VAE w/ NF, and \methodnameshort.
Because we keep the number of parameters comparable across the models \textit{at training}, the parameters of Causal VAE, Causal VAE w/ NF, and \methodnameshort \textit{at inference} are approximately 2$\times$ less than those of the AR baselines due to the dispensability of their encoders.

For further comparison with AR-Diffusion, we also train it with a backbone with half the number of Transformer blocks, so that EVA and AR-Diffusion have the same number of blocks at inference.

The models are trained for 400 epochs with AdamW optimizer~\cite{loshchilov2018decoupled} whose learning rate and $(\beta_1,\beta_2)$ are set to 1.0e-4 and (0.9, 0.95), respectively.

\noindent\textbf{Metrics.}
We report (FID)~\cite{heusel2017gans} and Inception Score (IS)~\cite{salimans2016improved} following the convention~\cite{mar}. 
For ImageNet we generate 50k samples while for VGGSound we generate 15446 (up to 50 samples per class, depending on the number of test samples).
We use the same number of test samples for target distributions for FID.
We additionally report the respective numbers of parameters at training and inference, excluding the tokenizer's parameters shared across models, and the inference time per single image generation normalized by ours.

\begin{figure}[t]
  \centering
  \includegraphics[width=\linewidth]{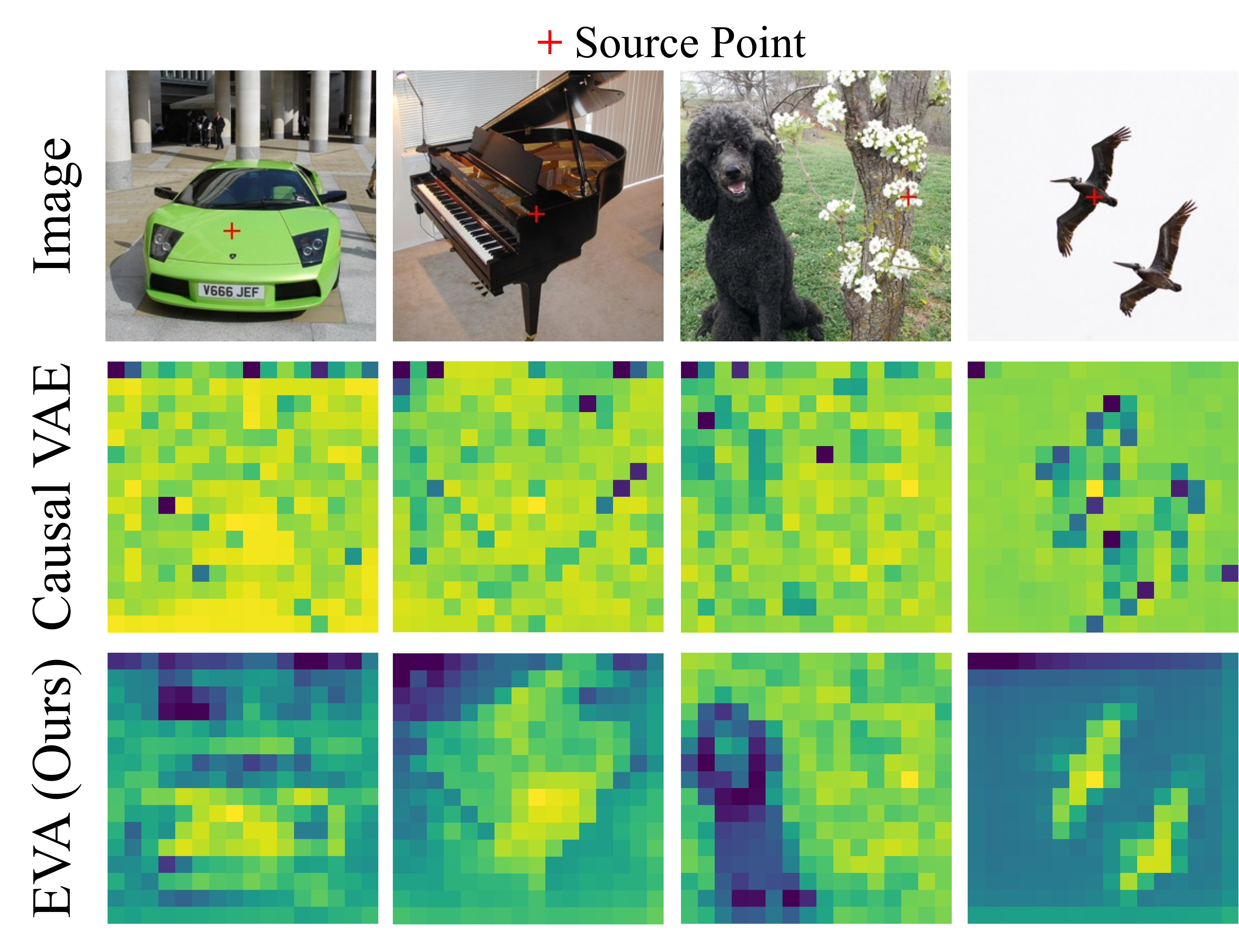}
    \caption{\textbf{Latent Space Visualization.} Each patch represents bi-directional KL divergence between the source latent ({\color{red}+}) and each latent. Lighter indicates similar to the source patch.}
    \label{fig:latent}
\end{figure}
\begin{figure}[t]
    \includegraphics[width=\linewidth]{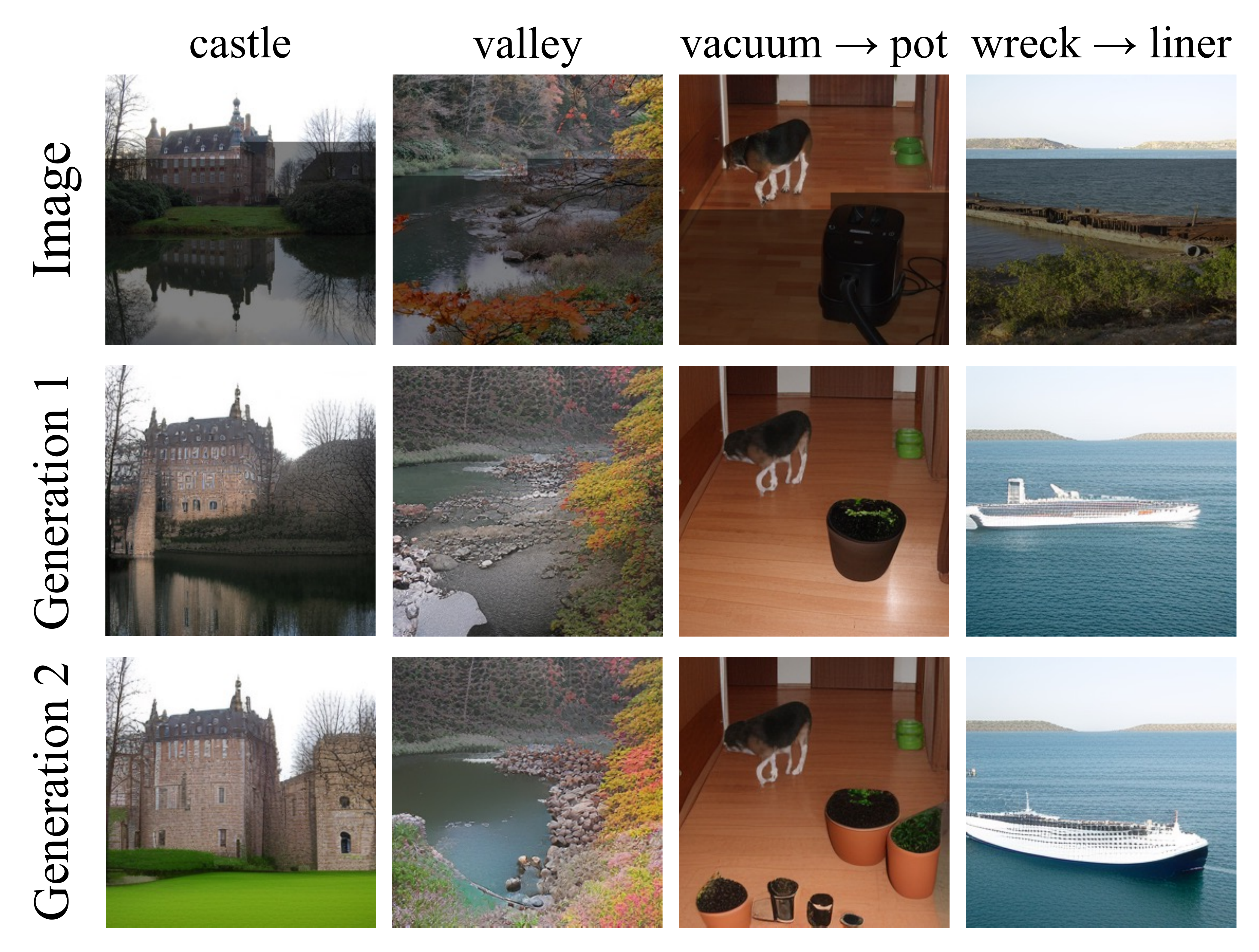}
    \caption{\textbf{Image Completion.} 
    The darker regions (like \gridthreefive \,) in the first row indicate the areas to be re-generated.}
    \label{fig:inpainting}
\end{figure}

\subsection{Comparison with Baselines}
Table~\ref{tab:result} shows the numerical comparisons on ImageNet and VGGSound.
Consistent with the toy study in Sec.~\ref{sec:toy}, the vanilla AR model trained with the MSE loss fails, indicating that direct regression is insufficient for multimodal next-token prediction in continuous space. 
Adding a GMM head~\cite{givt} substantially improves generation quality, but its finite mixture family remains restrictive. 

AR-Diffusion (321M), which has the same number of Transformer blocks at training time as EVA, achieves the best metrics but it requires the iterative denoising steps, at the cost of substantially slower inference (10 times slower than \methodnameshort with a comparable FID as shown in the right figure of Table~\ref{tab:result}). 
Moreover, we observe that AR-Diffusion (236M), which has the same number of blocks at inference time as EVA, remains at much worse FID. 
This result indicates that 
under matched Transformer depth at inference, \methodnameshort achieves substantially lower FID with fewer inference parameters.

Causal VAE performs poorly, demonstrating that simply making a VAE decoder causal is not sufficient for high-quality sequential generation.
In contrast, \methodnameshort achieves the competitive FID among methods with comparable inference cost.
These results suggest that learning an autoregressive latent prior is critical for high-fidelity continuous autoregressive generation.
Curated images are shown in Fig.~\ref{fig:teaser}. 
We include uncurated images in Appendix and sounds in the project page.

\subsection{Framework Analysis.}

\noindent\textbf{Latent Space Visualization.}
We found that \methodnameshort obtains a well-structured latent space after training.
In Fig.~\ref{fig:latent}, we compute bi-directional KL divergence, \ie, $\text{KL}(p||q)+\text{KL}(q||p)$, between a source patch ({\color{red}+}) and other patches in the posterior latent space.
The lighter patches indicate being more similar to the source patch.
We can see 
1) that in \methodnameshort's latent space, the semantically similar patches tend to have similar latent distributions, implying that learning generation naturally encourages the model to learn semantics,
and 2) that 
Causal VAE still exhibits a few spatially structured variations in its posterior latent distributions despite regularization toward the standard Gaussian, indicating residual mismatch with the factorized prior used for generation.

\begin{table}[t]
\centering

\begin{adjustbox}{width=1.0\linewidth}
\begin{tabular}{l| cc | cc}
    & \multicolumn{2}{c|}{\scriptsize{ImageNet}} & \multicolumn{2}{c}{\scriptsize{VGGSound}} \\
    \multicolumn{1}{l|}{Method} & {FID$\downarrow$} & {rFID$\downarrow$} & {FID$\downarrow$} & {rFID$\downarrow$} \\
    \midrule
    Causal VAE & 100.27  & 2.49 & 1.84 & 0.63\\
    Causal VAE w/ NF & 12.43& 2.64 & 0.65 & 1.08\\
    \methodnameshort & 7.68 & 2.42 & 0.64 & 0.64\\
    \end{tabular}
    \end{adjustbox}
\caption{\textbf{Generation vs. Reconstruction.} 
    \methodnameshort outperforms the baselines in generation fidelity (FID) while maintaining reconstruction fidelity (rFID).
}
\label{tab:genvsrec}
\end{table} 

\noindent\textbf{Image Completion.}
Our model is capable of image completion as shown in Fig.~\ref{fig:inpainting}.
We here use the first $t$ posterior latents $z_{1:t}$ from an original image $\mathbf{x}$, \ie, $q(z_{1:t} \mid \mathbf{x})$, to generate subsequent latents $z_{t+1:T}$.
In the first two rows, we give the same class conditions to the model as the original images while in the last two rows, we give the different class conditions, \ie, from ``vacuum'' to ``pot'' and from ``wreck'' to ``liner''.
We observe that \methodnameshort generates diverse samples from the same prefix contexts while the completed patches are aligned with the class conditions.

\noindent\textbf{Generation vs. Reconstruction.}
Classical VAEs with a single global latent bottleneck are often characterized by a trade-off between reconstruction fidelity and prior regularization.
In the multi-latent setting, as in our work, however, Causal VAE and \methodnameshort achieve comparable reconstruction fidelity, as reflected by their similar rFID in Table~\ref{tab:genvsrec}, while \methodnameshort substantially improves generation FID.
This result shows that generation fidelity can be substantially improved without sacrificing reconstruction fidelity, suggesting that the two objectives need not be tightly coupled when sufficient latent capacity is available.
By learning an autoregressive prior that captures dependencies across latent positions, \methodnameshort preserves the reconstruction capability of the VAE while significantly improving prior sampling.

\begin{figure}[t]
    \centering

    \begin{minipage}[c]{1.0\linewidth}
        \centering

        \begin{subfigure}[c]{0.49\linewidth}
            \centering
            \includegraphics[width=\linewidth]{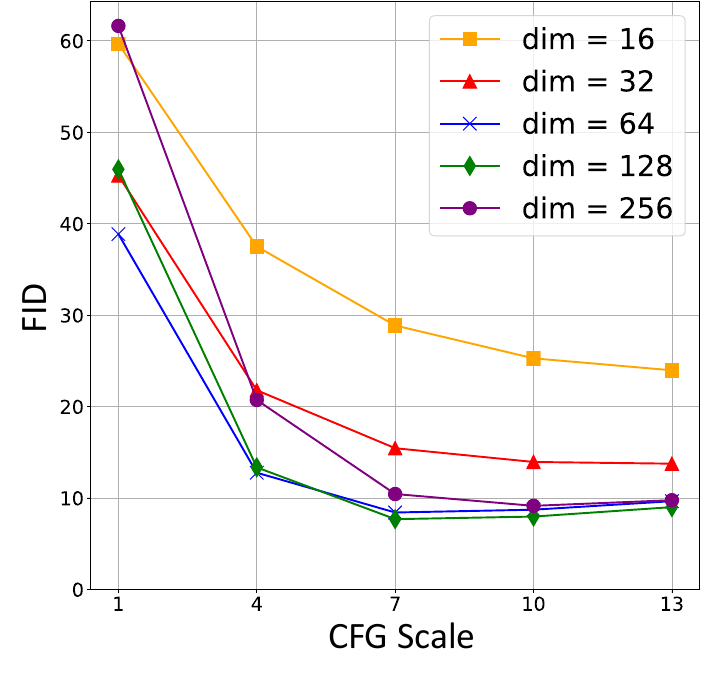}
            \vspace{-6mm}
            \caption{Latent Size.}
            \label{fig:dim}
        \end{subfigure}
        \hfill
        \begin{subfigure}[c]{0.49\linewidth}
            \centering
            \includegraphics[width=\linewidth]{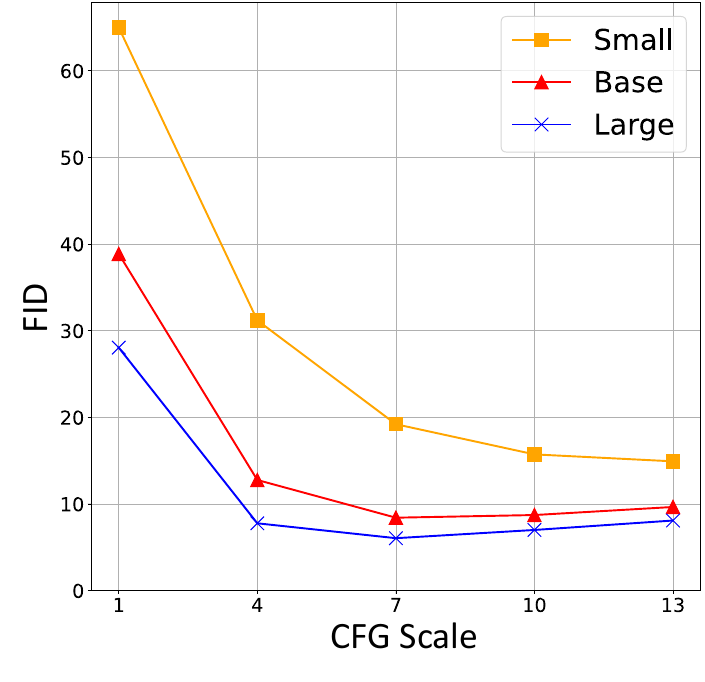}
            \vspace{-6mm}
            \caption{Model Size.}
            \label{fig:backbone}
        \end{subfigure}

    \end{minipage}
    \caption{
        \textbf{Latent Size and Model Size.} (a) Different latent sizes with EVA-B. The variant with a latent dimension of 128 achieves the best FID of 7.68 with a CFG scale of 7.
        (b) Different backbone sizes with a latent size of 64. 
        EVA-L achieves an FID of 6.04 with a CFG scale of 7.
    }
    \label{fig:factors}
\end{figure}

\noindent\textbf{Latent Size and Model Size.}
Fig.~\ref{fig:factors} analyzes two practical factors of \methodnameshort on ImageNet: the latent dimensionality and the backbone size over different CFG scales. 
In Fig.~\ref{fig:dim}, we vary the latent size from 16 to 256. 
A too-small latent bottleneck degrades FID indicating insufficient capacity to represent multimodal predictive structure. 
Increasing the latent size improves generation quality up to 128 dimensions, which gives the best FID on ImageNet. 
Further increasing the latent size does not yield additional gains and can even hurt performance, suggesting that excessively large latent spaces make prior modeling and sampling more difficult. 

We next study the effect of the model size in Fig.~\ref{fig:backbone}. 
The small variant (denoted as \methodnameshort-S) has 16 Transformer blocks each for encoder and decoder with a hidden width of 512 while the large variant (denoted as \methodnameshort-L) has 32 Transformer blocks each for encoder and decoder with a hidden width of 1024. 
All variants have the latent size of 64.
We observe that larger backbones successfully improve FID; \methodnameshort-B achieves an FID of 8.41 with a CFG scale of 7 while \methodnameshort-L achieves an FID of 6.04.

\section{Limitations and Future Work}
We identify two main limitations.
First, \methodnameshort currently linearizes spatial latent maps into a 1D causal sequence and therefore does not explicitly exploit their 2D spatial structure. 
Designing autoregressive priors specialized for spatial data may further improve image generation.
Second, we focus on closed-set generation benchmarks on ImageNet and VGGSound, and moderate model scales in this paper. 
A natural next step is to validate the method on larger-scale and more diverse downstream tasks, such as text-to-image~\cite{rombach2022high}, text-to-speech~\cite{meng2025autoregressive}, and multimodal data generation~\cite{zhou2024transfusion}.

\section{Conclusion}
We presented \methodname (\methodnameshort) for efficient sequence generation with continuous-valued tokens.
Unlike traditional VAEs, \methodnameshort learns the latent prior empirically from data only with an additional single linear layer on top of VAEs, which addresses the distribution gap between the prior and posterior and leads to high-fidelity sampling.
We demonstrated on ImageNet and VGGSound that our method achieves competitive generation quality with efficient ancestral sampling against autoregressive diffusion models.
We hope this work encourages the research community to re-examine the potential of VAEs as standalone generative models.

\clearpage
\appendix
\onecolumn

\section{Additional Experiments}
\begin{figure*}[b]
  \centering
\includegraphics[width=0.48\linewidth]{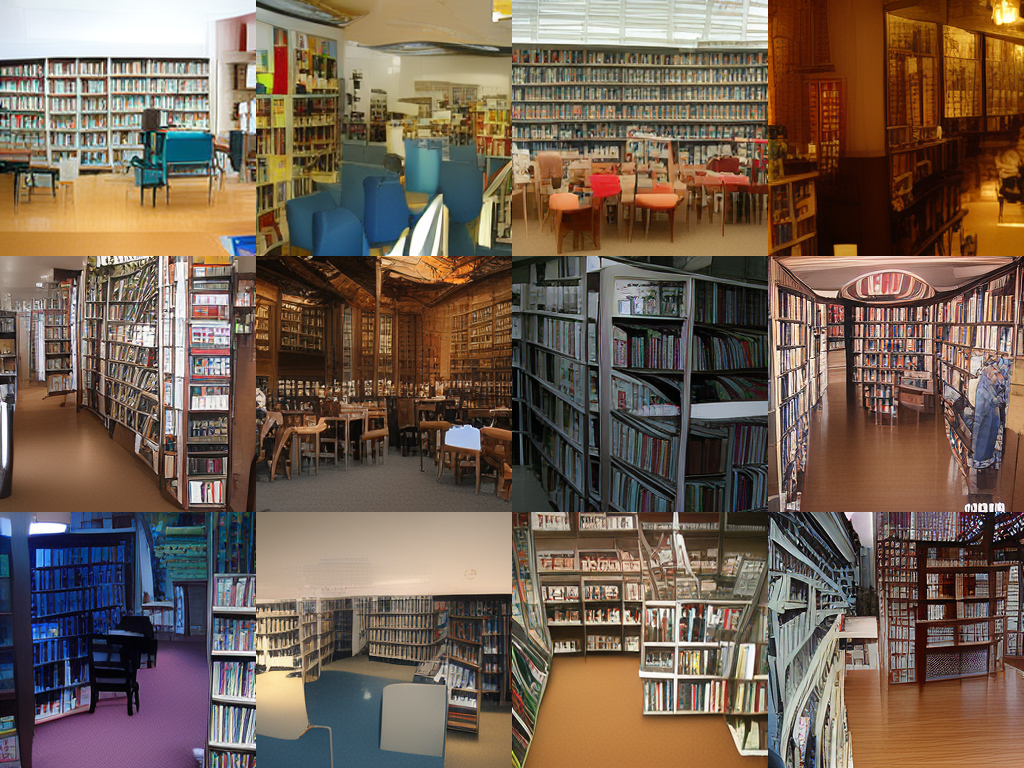}%
  \hfill
  \includegraphics[width=0.48\linewidth]{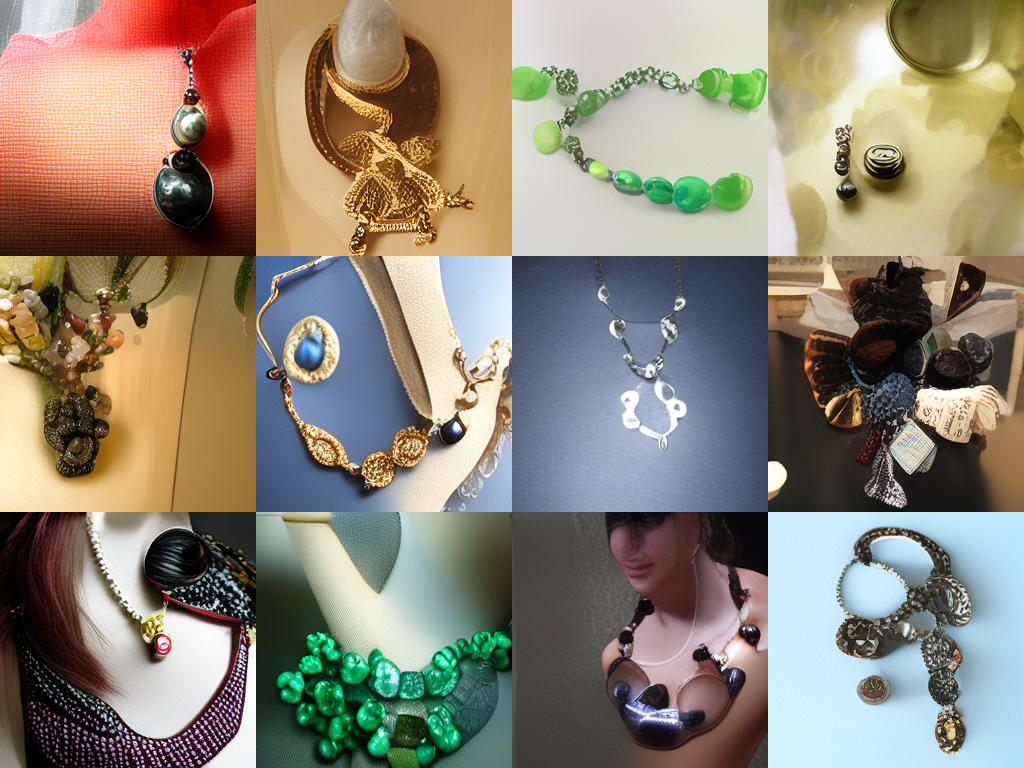}\\
  \vspace{1em}

  \includegraphics[width=0.48\linewidth]{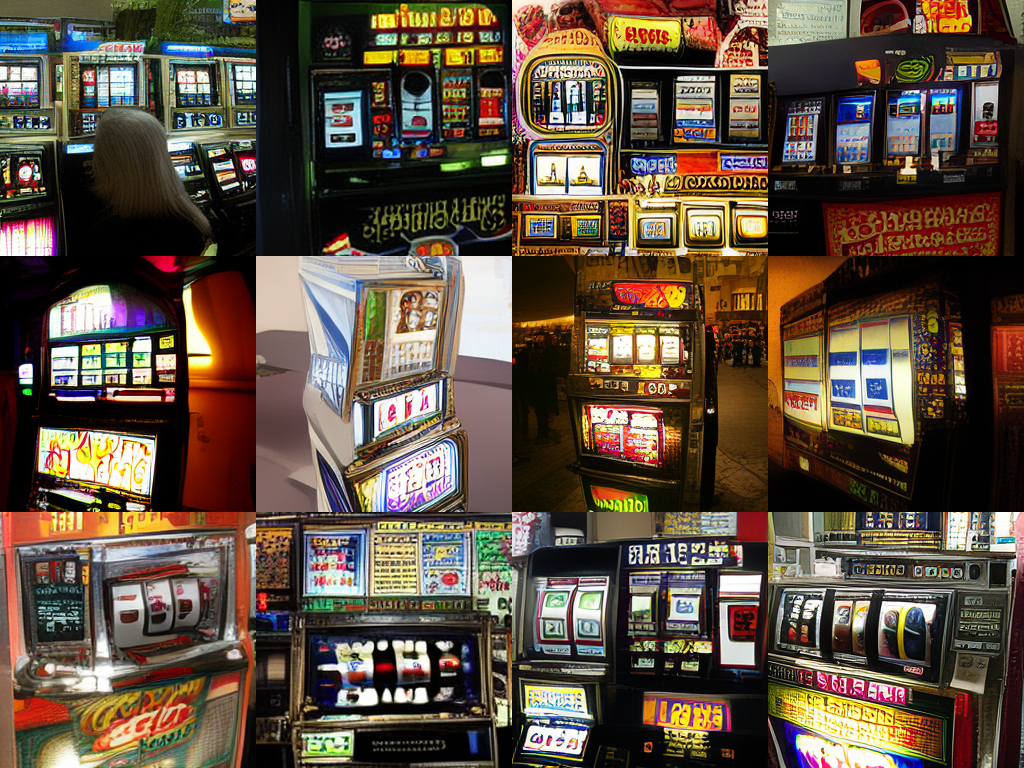}%
  \hfill
  \includegraphics[width=0.48\linewidth]{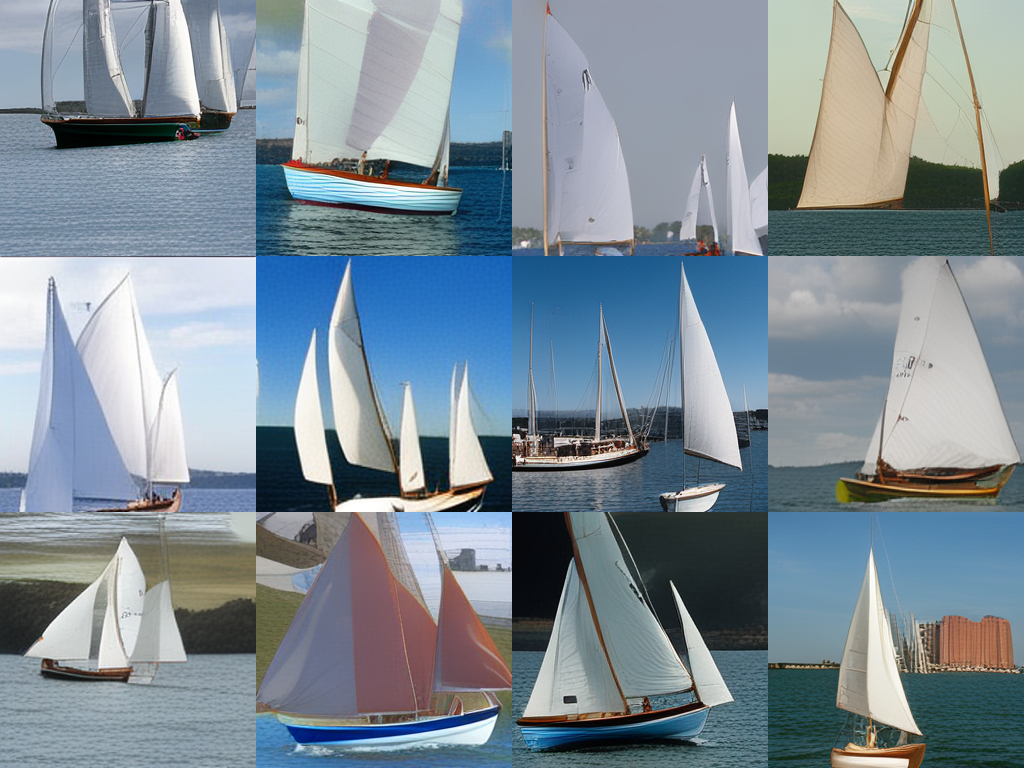}\\
  \vspace{1em}

  \includegraphics[width=0.48\linewidth]{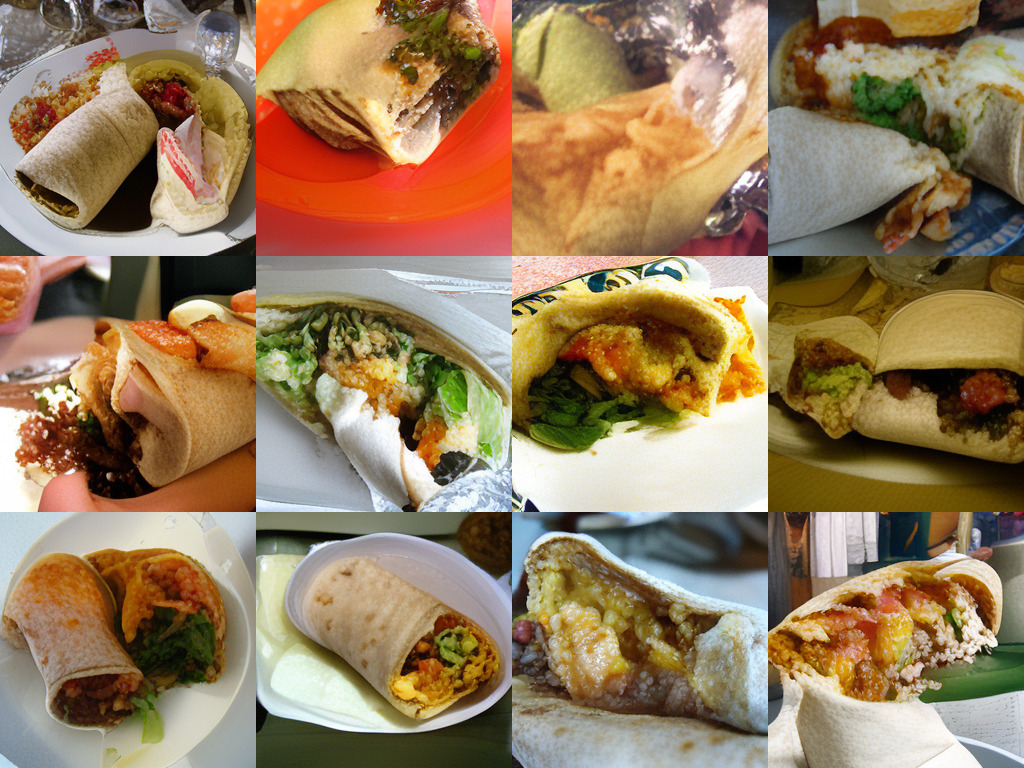}%
  \hfill
  \includegraphics[width=0.48\linewidth]{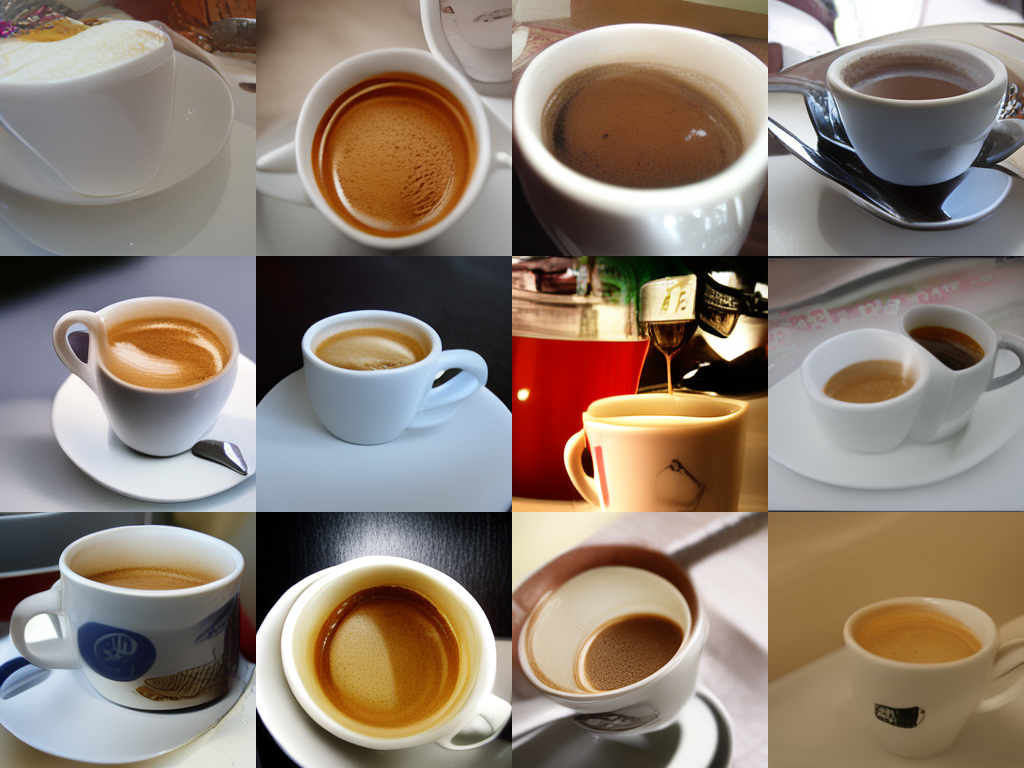}\\
  \vspace{0.5em}

  \caption{\textbf{Uncurated Examples.}}
  \label{fig:uncurated2}
\end{figure*}

\noindent
\begin{minipage}[t]{0.48\textwidth}
\vspace{0pt}
\noindent\textbf{System-level Comparison.}
We compare our method with recent state-of-the-art models, including VAR~\cite{var}, SiT~\cite{ma2024sit}, JiT~\cite{li2026back}, and LlamaGen~\cite{sun2024autoregressive}. 
We report inference parameters (including tokenizers), FID, and IS in Table~\ref{tab:sota}.
Note that bi-directional models such as VAR and SiT allow information exchange across the full spatial extent of the image representation at each generation stage, whereas causal models such as LlamaGen and our model condition each prediction only on previously generated tokens in raster-scan order.

\noindent\textbf{Uncurated Generated Examples.}
We show the uncurated images generated by EVA-L in Figs.~\ref{fig:uncurated}, and \ref{fig:uncurated2}.
Each block indicates the same class condition.
We also strongly recommend the readers listen to the generated sounds in the project page.

\end{minipage}
\hfill
\begin{minipage}[t]{0.48\textwidth}
\vspace{0pt}
\centering

\begin{adjustbox}{width=1.0\linewidth}
\begin{tabular}{l|c|cc}
    \multicolumn{1}{l|}{Method} & \#Params (M) & {FID$\downarrow$} & {IS$\uparrow$}\\
    \midrule
    {\color{gray}\textit{Bi-directional}} \\
    VAR-B & 197.9 & 12.82 & 126.34\\
    SiT-B & 179.9& 5.88 & 215.03\\
    JiT-B & 131.3 &3.72 & 267.87\\
    \midrule
    {\color{gray}\textit{Causal}} \\
    LlamaGen-B &153.4&8.68 & 120.37\\
    \methodnameshort-B & 124.6 &7.68 & 178.60 \\
    \end{tabular}
    \end{adjustbox}
\captionof{table}{\textbf{Comparison with the recent SOTA models on ImageNet-256.}}
\label{tab:sota}
% }
\end{minipage}

\vspace{4em}

\noindent
\begin{minipage}{\textwidth}
  \centering
  \includegraphics[width=0.48\linewidth]{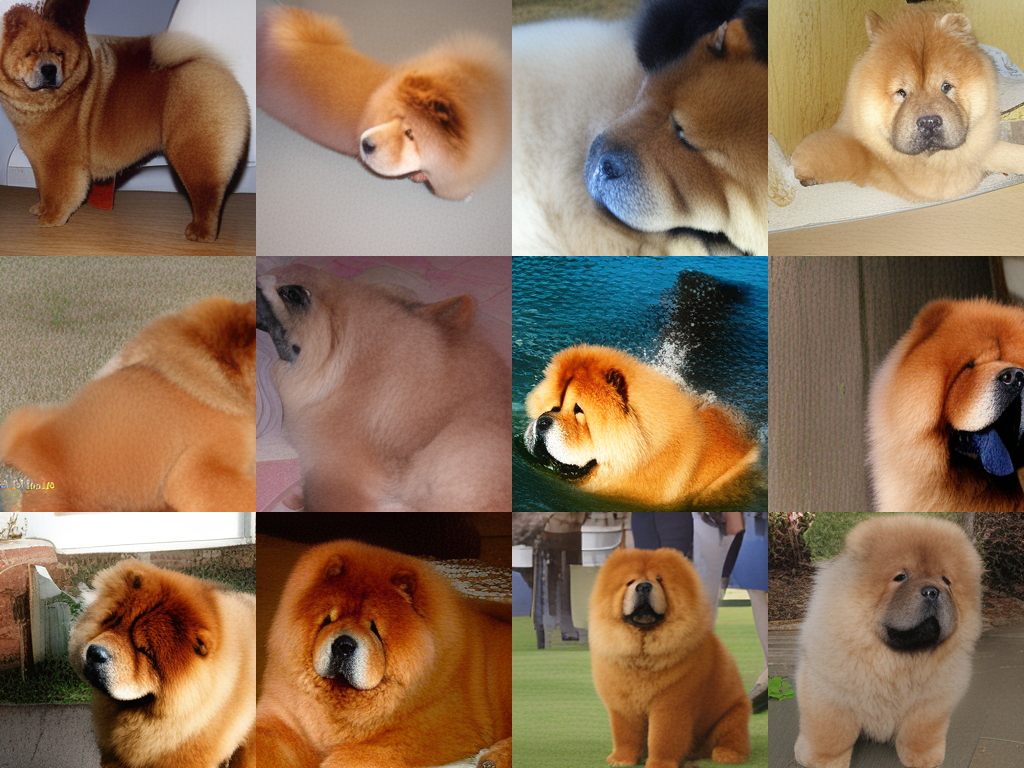}%
  \hfill
  \includegraphics[width=0.48\linewidth]{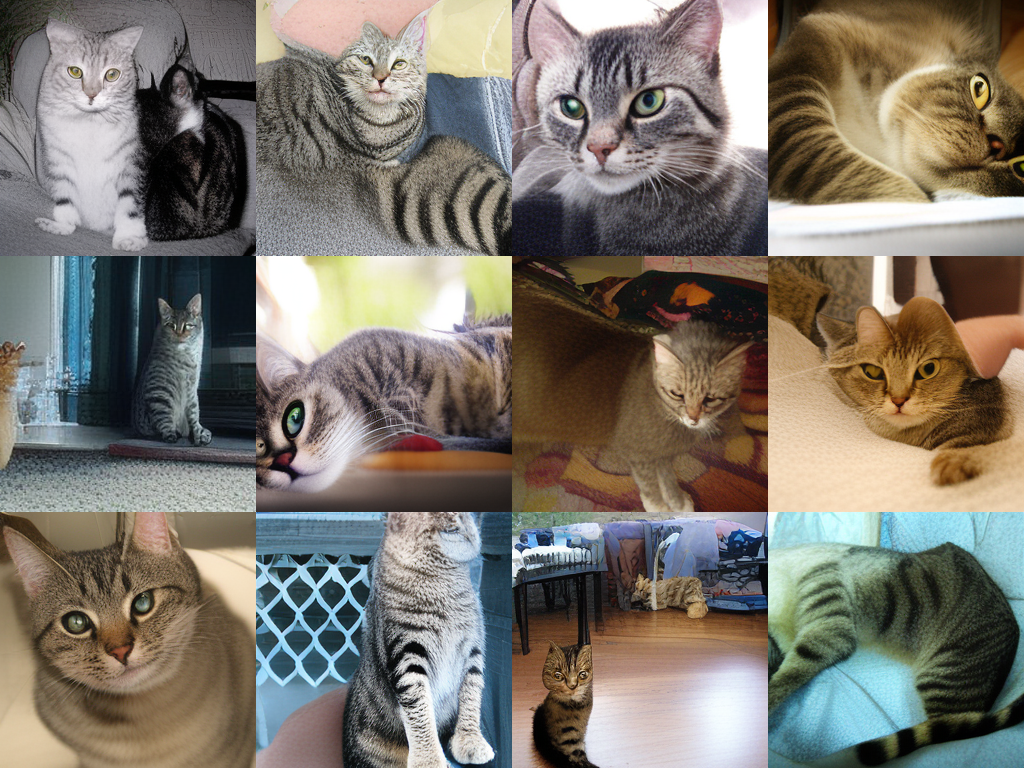}\\
  \vspace{1em}
  \includegraphics[width=0.48\linewidth]{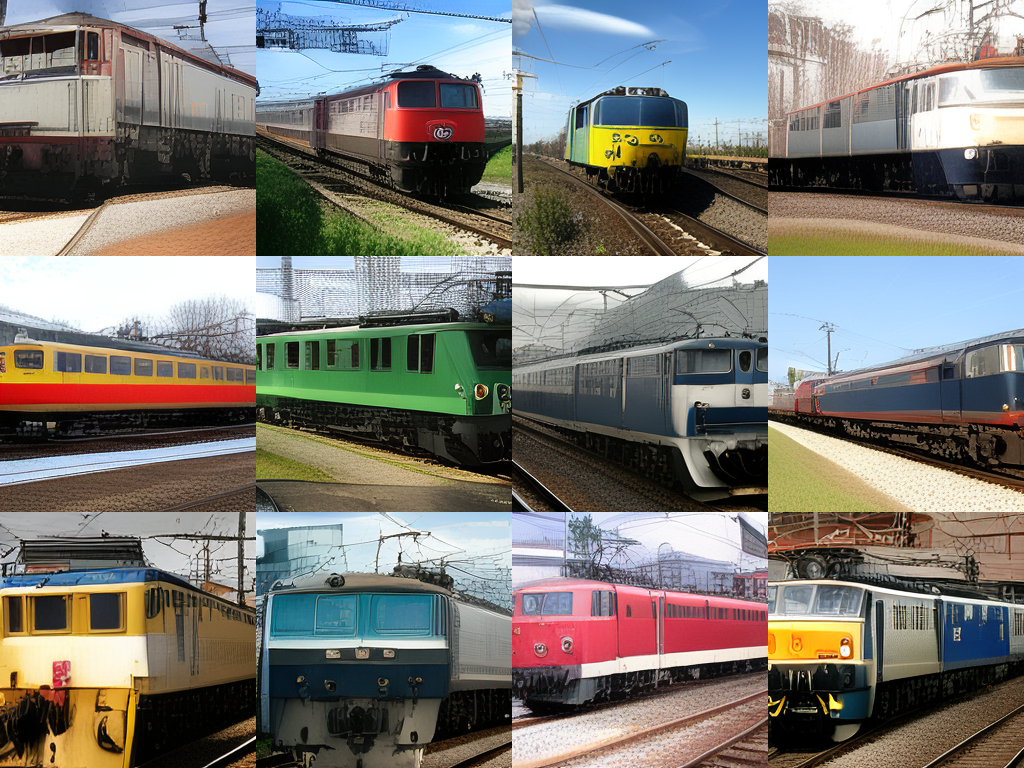}%
  \hfill
  \includegraphics[width=0.48\linewidth]{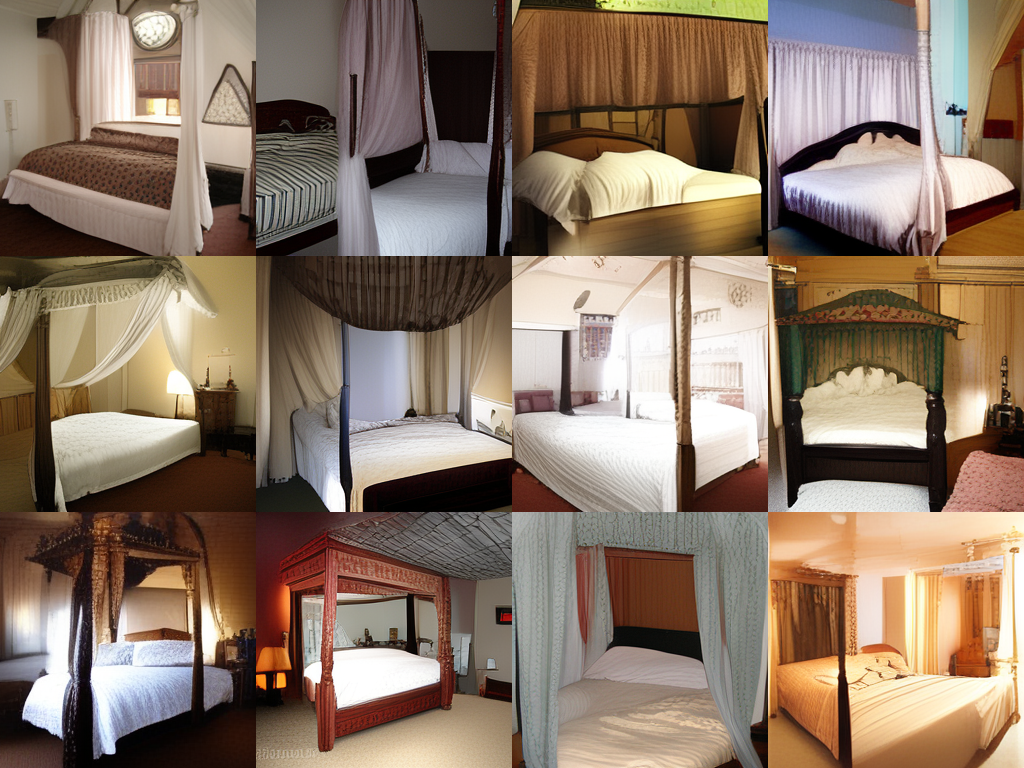}\\
  \vspace{0.5em}
  \captionof{figure}{\textbf{Uncurated Examples.}}
  \label{fig:uncurated}
\end{minipage}

\clearpage
\twocolumn

\clearpage

\section{Implementation Details}

\subsection{On the Toy Problem}
\label{sec:toy-detail}
\noindent\textbf{Data Generation.}
The toy benchmark is defined on a one-dimensional latent process and one-dimensional observations. Let $x_{\mathrm{BOS}}=0$ denote a fixed BOS token. We first sample the initial latent variable
\[
z_0 \sim \mathcal{N}(0, \sigma_{z_0}^2),
\]
with $\sigma_{z_0}=0.30$.
The next latent variable is generated through a state-dependent multimodal transition. Specifically, for a given number of latent modes $M_{\mathrm{lat}}$, we define mode logits
\[
\ell_k(z_0) = a_k z_0 + b_k, \qquad k=1,\dots,M_{\mathrm{lat}},
\]
where $a_k$ are linearly spaced between $-2.2$ and $2.2$, and
\[
b_k = 0.5 \cos\!\left(\frac{2\pi (k-1)}{M_{\mathrm{lat}}}\right).
\]
A latent mode is then sampled as
\[
m \sim \mathrm{Cat}(\mathrm{softmax}(\ell(z_0))).
\]
Given $m$, the next latent is generated by
\[
z_1 = 0.55 z_0 + 0.30 + c_m + \varepsilon_z,
\]
where $c_m$ is a mode-specific latent offset and $\varepsilon_z \sim \mathcal{N}(0, \sigma_{z_1}^2)$ with $\sigma_{z_1}=0.06$. The offsets $c_m$ are placed uniformly on a one-dimensional grid with spacing $1.6$.

Observations are generated from each latent state using observation-level multimodal offsets. For a given number of observation modes $M_{\mathrm{obs}}$, we define offsets $o_u$ on a one-dimensional grid with spacing $2.4$, and sample
\[
u_0, u_1 \sim \mathrm{Uniform}\{1,\dots,M_{\mathrm{obs}}\}.
\]
The observations are then generated as
\[
x_0 = z_0 + o_{u_0} + \varepsilon_0, \qquad
x_1 = z_1 + o_{u_1} + \varepsilon_1,
\]
where $\varepsilon_0,\varepsilon_1 \sim \mathcal{N}(0, \sigma_x^2)$ with $\sigma_x=0.05$.
In this construction, $M_{\mathrm{lat}}$ controls the number of branching transition modes in latent space, while $M_{\mathrm{obs}}$ controls the visible multimodality in observation space.

\noindent\textbf{AR and GMM Parameterizations.}
For the autoregressive baselines, we model
\[
p(x_0 \mid \mathrm{BOS}), \qquad p(x_1 \mid x_0, \mathrm{BOS}).
\]
In the deterministic AR baseline, each conditional distribution is parameterized by a separate multilayer perceptron (MLP). The network for $p(x_0 \mid \mathrm{BOS})$ takes a one-dimensional input and uses three hidden layers of width $128$ with ReLU activations, followed by a scalar output head. The network for $p(x_1 \mid x_0, \mathrm{BOS})$ takes a two-dimensional input $(x_{\mathrm{BOS}}, x_0)$ and uses the same hidden architecture.

For GMM, we use the same backbone MLPs, but replace the scalar output head with Gaussian mixture parameters. 
For each conditional distribution, the network outputs mixture logits, component means, and component log-standard-deviations for a $K$-component Gaussian mixture. 

\noindent\textbf{Causal VAE Parameterization.}
The VAE baseline models
\[
q(z_0, z_1 \mid x_0, x_1), \qquad
p(x_0 \mid z_0), \qquad
p(x_1 \mid z_0, z_1),
\]
with an unconditional prior
\[
p(z_0, z_1) = \mathcal{N}(0,I)\mathcal{N}(0,I).
\]
The encoder is an MLP that takes the two-dimensional input $(x_0,x_1)$ and uses three hidden layers of width $128$ with ReLU activations. It outputs the means and log-variances for both $z_0$ and $z_1$. 
We use latent dimension $8$ for each latent variable.
The decoder for $p(x_0 \mid z_0)$ is an MLP with two hidden layers of width $128$, mapping an $8$-dimensional latent input to a scalar output. 
The decoder for $p(x_1 \mid z_0, z_1)$ has the same hidden architecture and takes the concatenated $16$-dimensional latent vector $(z_0,z_1)$ as input.

The VAE is trained with the objective
\begin{align}
\mathcal{L}_{\mathrm{VAE}}
&=
\mathcal{L}_{\mathrm{rec}}(x_0)
+
\mathcal{L}_{\mathrm{rec}}(x_1)
\nonumber\\
&+
\mathrm{KL}(q(z_0\mid x_0,x_1)\,\|\,\mathcal{N}(0,I))
\nonumber\\
&+
\mathrm{KL}(q(z_1\mid x_0,x_1)\,\|\,\mathcal{N}(0,I))
,
\end{align}
where reconstruction terms are implemented as Gaussian negative log-likelihoods with fixed $\sigma_{\mathrm{rec}}=0.08$.

\noindent\textbf{\methodnameshort Parameterization.}
EVA uses the same encoder and decoder structure as the VAE, but replaces the unconditional latent prior with predictive latent priors
\[
r(z_0 \mid \mathrm{BOS}), \qquad r(z_1 \mid \mathrm{BOS}, z_0).
\]
The prior network for $r(z_0 \mid \mathrm{BOS})$ is an MLP with three hidden layers of width $128$ and ReLU activations, taking a one-dimensional BOS input and outputting the mean and log-variance of an $8$-dimensional Gaussian. 
The prior network for $r(z_1 \mid \mathrm{BOS}, z_0)$ uses the same hidden architecture, takes a $1+8=9$ dimensional input, and outputs the mean and log-variance of an $8$-dimensional Gaussian for $z_1$.

\methodnameshort is trained with
\begin{align}
\mathcal{L}_{\mathrm{EVA}}
&=
\mathcal{L}_{\mathrm{rec}}(x_0)
+
\mathcal{L}_{\mathrm{rec}}(x_1)
\nonumber\\
& +
\mathrm{KL}(q(z_0\mid x_0,x_1)\,\|\,r(z_0\mid \mathrm{BOS}))
\nonumber\\
& +
\mathrm{KL}(q(z_1\mid x_0,x_1)\,\|\,r(z_1\mid \mathrm{BOS}, z_0))
.
\end{align}
As in the VAE, we use $\sigma_{\mathrm{rec}}=0.08$.

\noindent\textbf{Training Details.}
All neural networks use ReLU activations. 
We train all models with Adam, learning rate $10^{-3}$, batch size $512$, and $20{,}000$ gradient steps. 
For VAE and \methodnameshort, latent variables are sampled using the reparameterization trick. 
During evaluation, we generate samples unconditionally by following each model's native generative pathway from the BOS token.

\subsection{On the ImageNet and VGGSound}

\noindent\textbf{Transformer Backbone for Causal VAE and \methodnameshort.}
Both encoder and decoder transformers operate at a hidden width of 768 with 12 attention heads and 12 Transformer blocks each, using an MLP expansion ratio of 4.0 and LayerNorm~\cite{layernorm}.
Attention and projection dropouts are set to 0.1, and QKV projections include bias terms. 
The model prepends a single class token and additional four register tokens to the token sequence. 
On the encoder side, token embeddings are projected from the VAE latent token space into the 768-dimensional Transformer space, and the encoder applies full-attention. 
On the decoder side, attention is strictly causal, and latent variables are projected into the same 768-dimensional space before autoregressive decoding. 
RoPE~\cite{rope} is used in both encoder and decoder.

\noindent\textbf{Transformer Backbone for AR Baselines.}
We use a causal Transformer backbone with 24 blocks to keep the training parameters comparable with Causal VAE and \methodnameshort.
Other configurations follow Causal VAE and \methodname.

\noindent\textbf{AR-GMM.}
We follow GIVT~\cite{givt} for the implementation of the GMM head.
Each autoregressive Transformer state is projected by a linear head to the parameters of a Gaussian mixture distribution for the next latent token, namely mixture weights and component-wise means and scales. 
The scale parameters are transformed by a square-plus function and lower-bounded by $10^{-6}$ to ensure positivity and numerical stability. 
The model is trained by minimizing the negative log-likelihood of this mixture distribution.

At inference time, we follow the same sampling scheme as GIVT and use a temperature of $1.0$ for the mixture weights and $1.0$ for the component scales. 
Under classifier-free guidance, we adopt the same density combination rule as GIVT and perform rejection sampling with 1000 proposals.

\noindent\textbf{AR-Diffusion.}
We follow MAR~\cite{mar} for the implementation of the diffusion head.
Concretely, an AdaLN-conditioned MLP~\cite{dit} is employed as the diffusion head on top of the autoregressive transformer. 
The conditioning dimension is 768, the MLP hidden width is 1536, and the head depth is 12 residual blocks. 
The MLP jointly predicts the mean-related (noise) term and the variance term in the diffusion parameterization. 
Temporal conditioning uses a sinusoidal embedding of size 256 followed by a 2-layer MLP, which is combined additively with the AR conditioning signal and injected through AdaLN in each residual block. 
Training uses a cosine diffusion noise schedule. 
Differently from the original implementation~\cite{mar}, we use the DDIM~\cite{ddim} sampler instead of DDPM~\cite{ddpm} during inference because of its efficiency and fidelity.

\noindent\textbf{Causal VAE w/ NF.}
We follow SimFlow~\cite{zhao2025simflow} to implement normalizing flow (NF) prior.
It has additional 88.1M parameters consisting of six transformations, each parameterized by a two-layer Transformer with hidden dimension 768, 12 attention heads, and an MLP expansion ratio of 4.
The affine output layer predicts 256-dimensional log-scales and shifts. 
We train the prior by minimizing the change-of-variables negative log-likelihood of posterior samples under a standard Gaussian base distribution, jointly with the latent reconstruction objective.

\noindent\textbf{Additional Training Details.}
Optimization follows AdamW with a weight decay of 0.02. 
We use linear learning-rate warmup for the first 50 epochs.
For stability, we maintain an exponential moving average (EMA) of model weights, updated after each optimizer step with a decay of 0.9999.
We drop the class condition with a probability of 0.1 and insert a learnable unconditional token for CFG.
The training takes about 3 days for ImageNet and 12 hours for VGGSound with eight NVIDIA RTX5090 (32GB) GPUs.


% \clearpage
{
    \small
    \bibliographystyle{ieeenat_fullname}
    \bibliography{main}
}

\end{document}